\def\year{2020}\relax
\documentclass[letterpaper]{article} % DO NOT CHANGE THIS
\usepackage{aaai20}  % DO NOT CHANGE THIS
\usepackage{times}  % DO NOT CHANGE THIS
\usepackage{helvet} % DO NOT CHANGE THIS
\usepackage{courier}  % DO NOT CHANGE THIS
\usepackage[hyphens]{url}  % DO NOT CHANGE THIS
\usepackage{graphicx} % DO NOT CHANGE THIS
\usepackage{graphicx}  % DO NOT CHANGE THIS
\usepackage{bm}
\usepackage{multirow}
\usepackage{epstopdf}
\usepackage{subfigure}
\title{Shallow Feature Based Dense Attention Network for Crowd Counting}

\author{
\textbf{Yunqi Miao,\textsuperscript{1} Zijia Lin,\textsuperscript{2} Guiguang Ding,\textsuperscript{3} Jungong Han\textsuperscript{1}\thanks{Corresponding Author}}\\ 
\textsuperscript{1} University of Warwick, Coventry, UK \\
\textsuperscript{2} Microsoft Research, Beijing, China  \\
\textsuperscript{3} Tsinghua University, Beijing, China  \\
\textsuperscript{1} \{Yunqi.Miao.1, Jungong.Han\}@warwick.ac.uk \\
\textsuperscript{2} zijlin@microsoft.com	\\
\textsuperscript{3} dinggg@tsinghua.edu.cn \\
}

\newcommand{\etc}{\textit{etc}}
\newcommand{\etal}{\textit{et al.}}
\newcommand{\eg}{\textit{e.g.}}
\newcommand{\ie}{\textit{i.e.}}

\newcommand{\SDANet}{SDANet}

\begin{document}
\maketitle
\begin{abstract}
While the performance of crowd counting via deep learning has been improved dramatically in the recent years, it remains an ingrained problem due to cluttered backgrounds and varying scales of people within an image. In this paper, we propose a \textbf{S}hallow feature based \textbf{D}ense \textbf{A}ttention \textbf{Net}work (\SDANet{}) for crowd counting from still images, which diminishes the impact of backgrounds via involving a shallow feature based attention model, and meanwhile, captures multi-scale information via densely connecting hierarchical image features. Specifically, inspired by the observation that backgrounds and human crowds generally have noticeably different responses in shallow features, we decide to build our attention model upon shallow-feature maps, which results in accurate background-pixel detection. Moreover, considering that the most representative features of people across different scales can appear in different layers of a feature extraction network, to better keep them all, we propose to densely connect hierarchical image features of different layers and subsequently encode them for estimating crowd density. Experimental results on three benchmark datasets clearly demonstrate the superiority of \SDANet{} when dealing with different scenarios. Particularly, on the challenging UCF\_CC\_50 dataset, our method outperforms other existing methods by a large margin, as is evident from a remarkable 11.9$\%$ Mean Absolute Error (MAE) drop of our SDANet.
\end{abstract}

\section{Introduction}

\noindent Crowd counting aims to count the number of people by means of estimating the density distribution of the crowd in a single image. It is a very useful computer vision technique to facilitate a variety of applications, including crowd control, disaster management and public safety monitoring. However, it is not a trivial task due to great challenges in real-world situations caused by cluttered backgrounds and non-uniform people scale within an image. 

\begin{figure}
	\centering
	\subfigure[Background noise]{
		\label{Crowd.1}
		\includegraphics[width=0.3\textwidth]{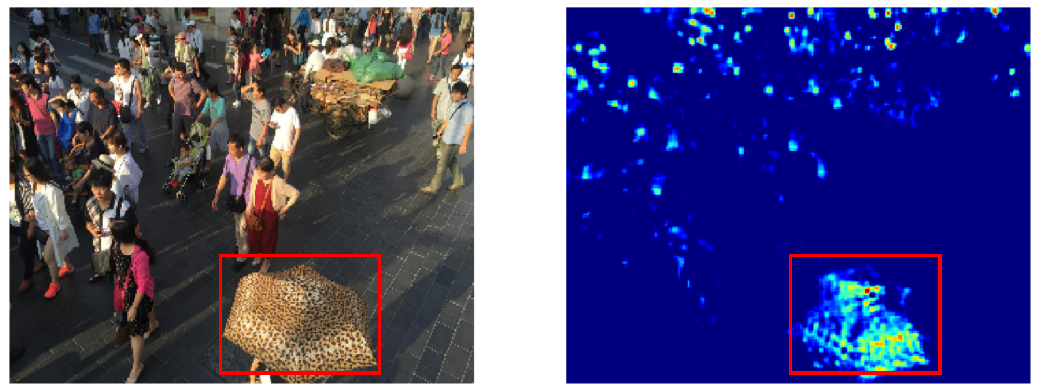}}
	\subfigure[Scale variation]{
		\label{Crowd.2}
		\includegraphics[width=0.15\textwidth]{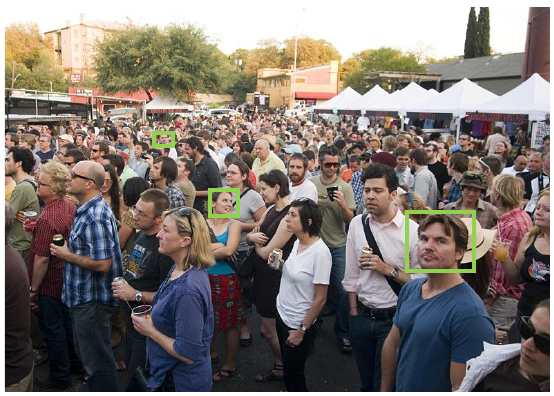}}
	\caption{Illustrations of the problems of cluttered backgrounds and varying scales of people. In (a), the right picture depicts the estimated density map of the left image, where backgrounds like the umbrella (in red box) could be mistakenly regarded as people in the density map and thus decrease the estimation accuracy. In (b), sizes of human heads (in green boxes) vary greatly within the image due to their different distances from the camera.}
	\label{fig:1}
\end{figure}

Tremendous algorithms \cite{zhang2016single,li2018csrnet,jiang2019crowd} have been proposed in the literature for estimating the crowd density distribution. The majority of them focused on addressing two problems when learning the mappings from image features to density distribution maps, \ie, 1) how to eliminate the impacts of cluttered backgrounds, and 2) how to deal with varying scales of people within an image. Figure\ \ref{fig:1} illustrates both mentioned problems. Specifically, in Figure\ \ref{Crowd.1}, the right picture depicts the estimated density map of the left image, derived by the MCNN model \cite{zhang2016single}. It can be noticed that backgrounds, \eg, umbrellas, could be mistakenly regarded as people on the density map, thus decreasing the estimation accuracy. Meanwhile, as illustrated in Figure\ \ref{Crowd.2}, sizes of human heads can vary greatly within an image, because of their different distances from the camera.

To eliminate noises caused by cluttered backgrounds, attention mechanism is usually introduced to re-weigh features or regions in terms of their probabilities of being the crowds. Generally, additional training samples and parameters are employed to train standalone classifiers indicating density levels \cite{sam2017switching} or head probability \cite{liu2019adcrowdnet} as the metric to evaluate the importance of different features/regions within an image, on the basis of which, different weights are given to the features/regions. However, standalone networks with complex structures usually require millions of extra to-be-learned parameters, which can be a heavy burden for a real-life application. 

\begin{figure}
	\centering
	\subfigure[]{
		\label{MCNN.1}
		\includegraphics[width=0.4\textwidth]{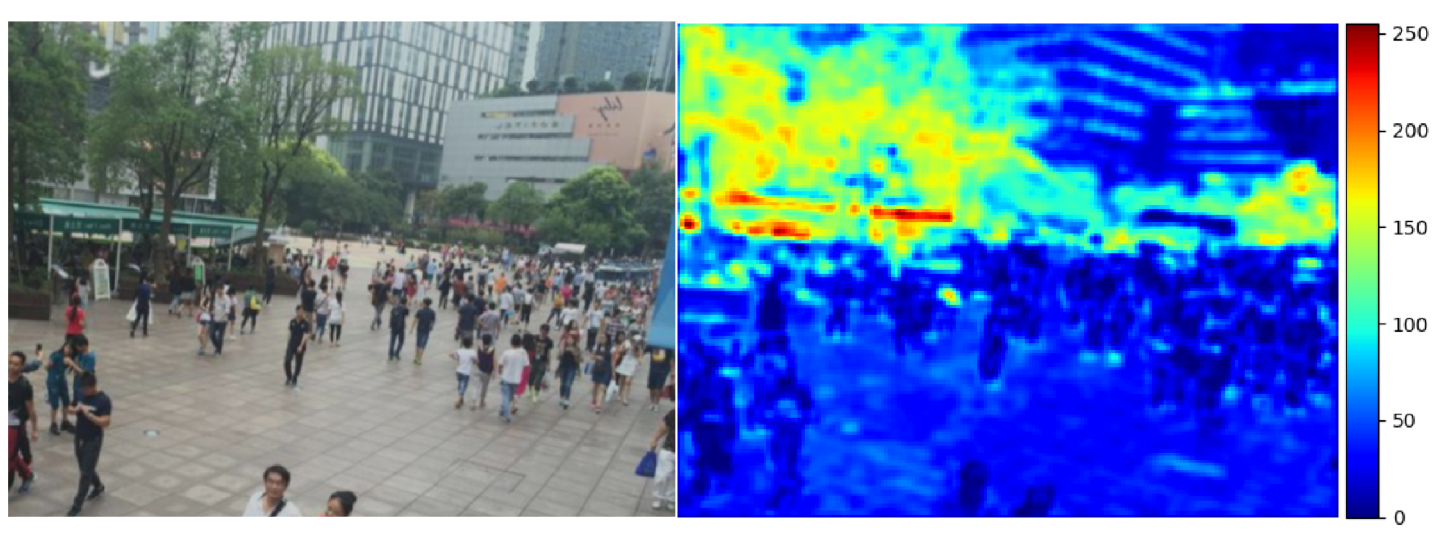}}
	\subfigure[]{
		\label{MCNN.2}
		\includegraphics[width=0.4\textwidth]{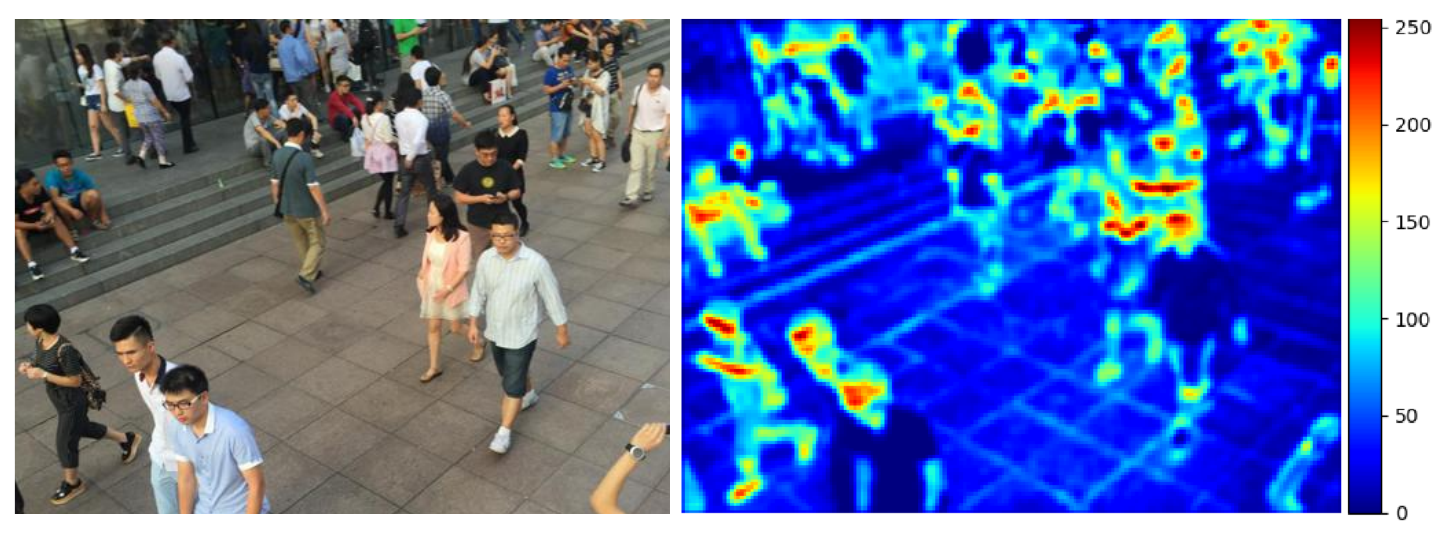}}
	\caption{Images and their corresponding shallow feature maps from several baselines. The shallow feature maps are linearly normalized to [0,255] by their maximums, which are shown as heat maps. It can be seen that the backgrounds and the human crowds have significantly different responses in (a) and (b).}
	\label{fig:2}
\end{figure}

By exploring the relationship between images and their corresponding normalized shallow feature maps generated by several baselines \cite{zhang2016single,boominathan2016crowdnet,li2018csrnet} (Figure\ \ref{fig:2}), we observe, for the first time, that backgrounds like stairs, trees and buildings, tend to have significantly different responses from those of the human crowds. For example, the backgrounds have stronger responses in Figure\ \ref{MCNN.1} but weaker ones in Figure\ \ref{MCNN.2}, whereas human crowds' reactions are opposite (weaker responses in Figure\ \ref{MCNN.1} but stronger ones in Figure\ \ref{MCNN.2}). This tells us that backgrounds and human crowds are more separable on shallow-layer feature maps. An attention model based on shallow features has potential to generate more accurate attention maps. Therefore, instead of involving a sophisticated standalone attention model as previous works, we incorporate an attention module in our feature extraction networks, which effectively reuses the shallow features and enjoys less complex structures to diminish background noises.

Regarding the problem of varying scales of people within an image, some works \cite{zhang2016single,deb2018aggregated} adopted ``multi-column'' frameworks to extract multi-scale information from images, where each branch extracts features of a specific scale by adopting filters with a certain size. Others exploit some convolutional operations, like dilated \cite{li2018csrnet,deb2018aggregated} and deformable convolution kernels \cite{liu2019adcrowdnet,zou2018danet}, to capture multi-scale information by expanding the receptive field of filters. Yet most of them extracted features layer by layer, and thus the features of the current layer may lose information of features in some preceding layers.

Actually, the most representative features of people across different scales can appear in different layers of the feature extraction networks. For example, the most representative features of people in a smaller scale can probably be extracted in an earlier layer, while those of people in a larger scale can be extracted in a later layer. Thus, it is vital to keep information of features in all different layers. Therefore, densely-connected structure that enables each layer to process features from all preceding layers seems like an appropriate structure, on which features corresponding to all scales can be well preserved and better encoded to facilitate the estimation of the crowd density. 

Based on the observations above, we propose a new method for crowd counting, termed \textbf{S}hallow feature based \textbf{D}ense \textbf{A}ttention \textbf{Net}work (\SDANet{}). \SDANet{} consists of three components, \ie, low-level feature extractor, high-level feature encoder, and attention map generator. As mentioned above, the attention map generator reduces the noises caused by backgrounds via re-weighing specific regions with attention maps generated with shallow features. Moreover, multi-scale information is well preserved via densely connecting the features of different layers in the high-level feature encoder. Extensive experiments on benchmark datasets also clearly demonstrate the superiority of \SDANet{}.

Contributions of our work are summarized as follows:
\begin{itemize}
\item 
We observe, for the first time, that shallow features contain distinguishable information between backgrounds and human crowds, which allows us to utilize a lightweight network to generate even more accurate attention maps.
\item 
We propose to employ densely connected structures in feature extraction/encoding networks, such that multi-scale information in different layers can be well kept to facilitate the estimation of the crowd density.
\item 
We propose a novel crowd counting method termed \SDANet{}. And experiments conducted on three benchmark datasets show that \SDANet{} achieves the state-of-the-art performance for crowd counting.
\end{itemize}

\begin{figure*}
	\centering  
	\subfigure[SDANet]{
		\label{SDANet}
		\includegraphics[width=0.9\textwidth]{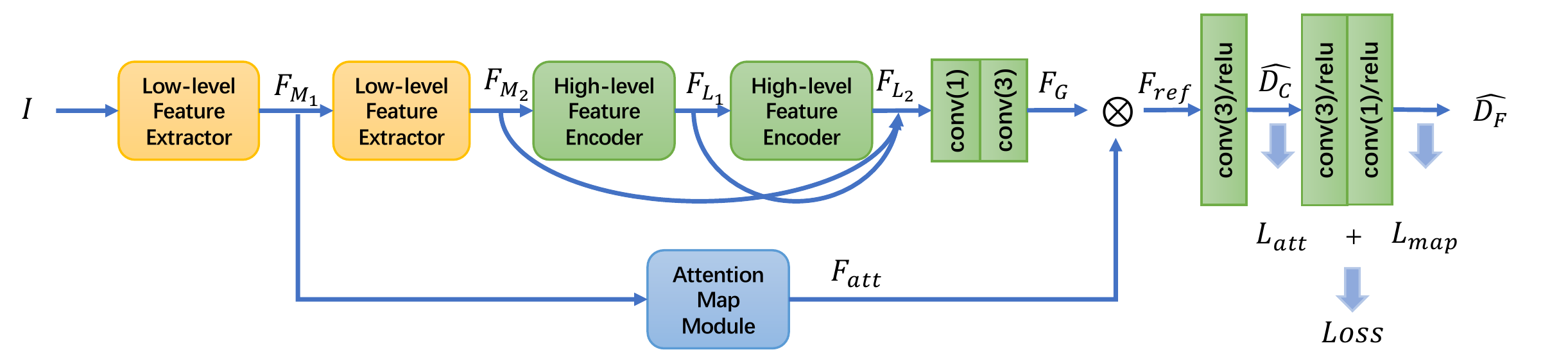}}
	\subfigure[HFE]{
		\label{HFE}
		\includegraphics[width=0.4\textwidth]{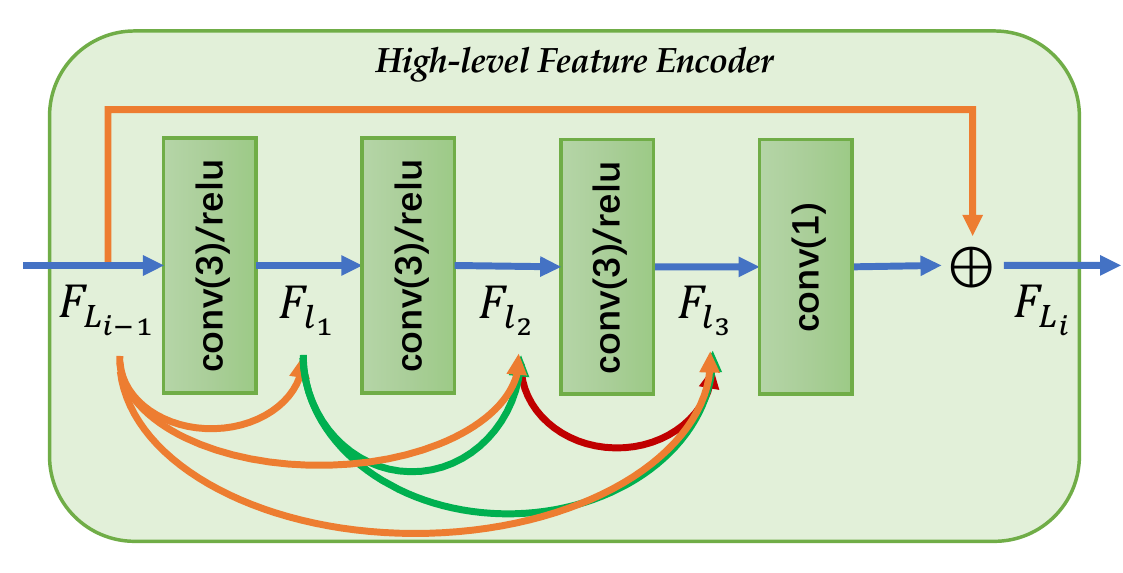}}
	\subfigure[AMG]{
		\label{AMG}
		\includegraphics[width=0.3\textwidth]{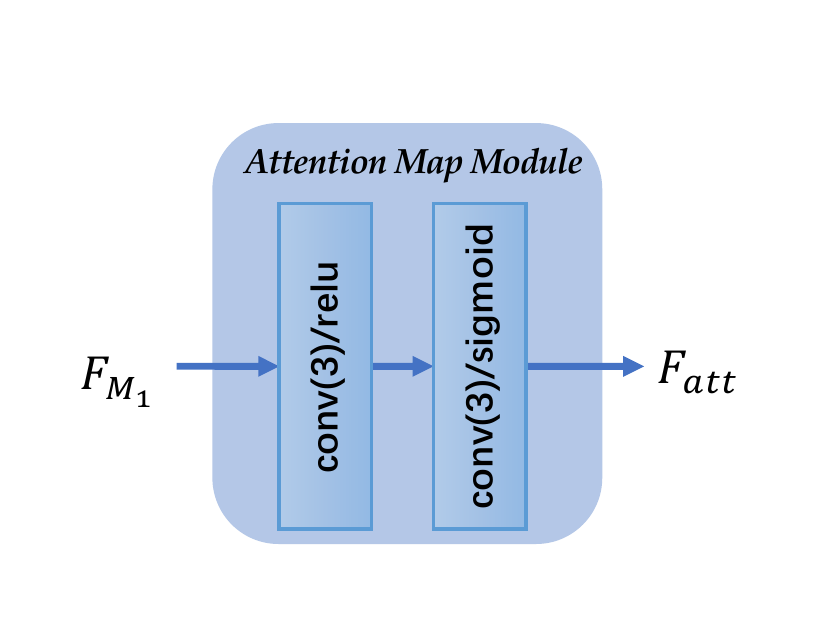}}
	\caption{ (a) The architecture of \SDANet{}. (b) The architecture of HFE. (c) The architecture of AMG.}
	\label{fig:3}
\end{figure*}

\section{Related Works}
Over the last few years, researchers have attempted to address the issue of crowd counting by density estimation with a variety of approaches \cite{sindagi2018survey}, where a mapping from image features to crowd density is learned and then the counted number is the summation over an estimated density map. Existing density estimation methods can be generally categorized as hand-crafted feature based ones and deep feature based ones, where latter ones tend to incorporate attention mechanism recently. 

\subsection{Hand-Crafted Feature based Methods}
Early works usually extract hand-crafted features implying global image characteristics, such as local binary pattern (LBP) and gray level co-occurrence matrices (GLCM), and learn its mapping to the density by regression models, ranging from linear ones to non-linear ones. Lempitsky \etal \cite{lempitsky2010learning} utilized linear models to describe the mapping from image features to the density in a local region, which is applied in bacteria counting and crowd counting with a relatively sparse density. Idress \etal \cite{idrees2013multi} explored features from three sources, \ie, Fourier, interest points and head detection combined with their respective confidences to get counts at localized patches and adopted a Markov Random Field (MRF) framework to obtain an estimated count for the entire image. 

\subsection{Deep Feature based Methods}
Inspired by the huge success of convolutional neural networks (CNN) in image classification \cite{krizhevsky2012imagenet}, recently deep features have been leveraged for density estimation. Owing to their superior performance, deep learning based methods \cite{wang2015deep,zhang2016single,deb2018aggregated,li2018csrnet,hossain2019crowd} quickly dominate the research in crowd counting. 

Zhang \etal \cite{zhang2016single} proposed a multi-column based architecture (MCNN), where each column adopts a filter with a certain size to extract features of the corresponding scale. Instead of training all patches with the same paralleled network, Sam \etal \cite{sam2017switching} proposed a switching CNN that adaptively selects the optimal branch for an image patch according to its density. A classifier indicating patch density is trained beforehand and empowers density estimation networks by providing prior knowledge. Recently, dilated kernels have also been involved in multi-column frameworks to further deliver larger reception fields \cite{li2018csrnet}.

\subsubsection{Attention mechanism in crowd counting}
Recently, attention mechanism is widely incorporated to enhance the crowd counting performance.
The idea is to roughly approximate the regions in the image where people are likely appeared. To do so, an attention model is learned to assign larger weights to pixels/regions of being human crowds. \cite{liu2018decidenet,kang2018crowd,hossain2019crowd,liu2019adcrowdnet,zhu2019dual}. 

ADCrowdNet \cite{liu2019adcrowdnet} employs an attention map generator trained on additional negative samples and then applies it to detect crowd regions in the images. Hossain \etal \cite{hossain2019crowd} proposed a Scale-Aware Attention Networks (SAAN), which utilizes attention mechanism to re-weigh multi-scale features learned by multi-columns. SFANet \cite{zhu2019dual} generates an attention map with the same size of the image by an additional CNN branch, where each pixel indicates its probability of being the head. Alternatively, DecideNet \cite{liu2018decidenet} uses a learned attention map to combine the two maps generated by the regression branch and the detection branch. 

The proposed \SDANet{} in this paper is also a deep feature based method with attention mechanism incorporated. However, different from previous works that learned a standalone attention model with sophisticated structures, by observing that shallow features can have strong signals to distinguish backgrounds and human crowds, we propose to use shallow features to build an attention module in \SDANet{} with simpler network structures. Moreover, instead of encoding multi-scale features layer by layer that has the risk of losing feature information of some preceding layers, we propose to densely connect outputs of each layer in \SDANet{}, so that multi-scale features of different layers can be better kept and encoded to facilitate the estimation of crowd density.

\section{Our Approach}
The framework of \SDANet{} is illustrated schematically in Figure\ \ref{SDANet}, which mainly consists of three components: Low-level Feature Extractor (LFE), High-level Feature Encoder (HFE), and Attention Map Generator (AMG). 

\subsection{Low-level Feature Extractor (LFE)}
Most existing methods use separate branches with different size filters to extract multi-scale information from images, which may introduce redundant structures into the pipeline \cite{li2018csrnet}. Inspired by the success of SANet \cite{cao2018scale} in feature extraction, the Inception module \cite{szegedy2015going}, a tool to process visual information of various scales, is used as the shallow feature extractor of \SDANet{}. 

\begin{figure}
	\centering
	\includegraphics[width=0.45\textwidth]{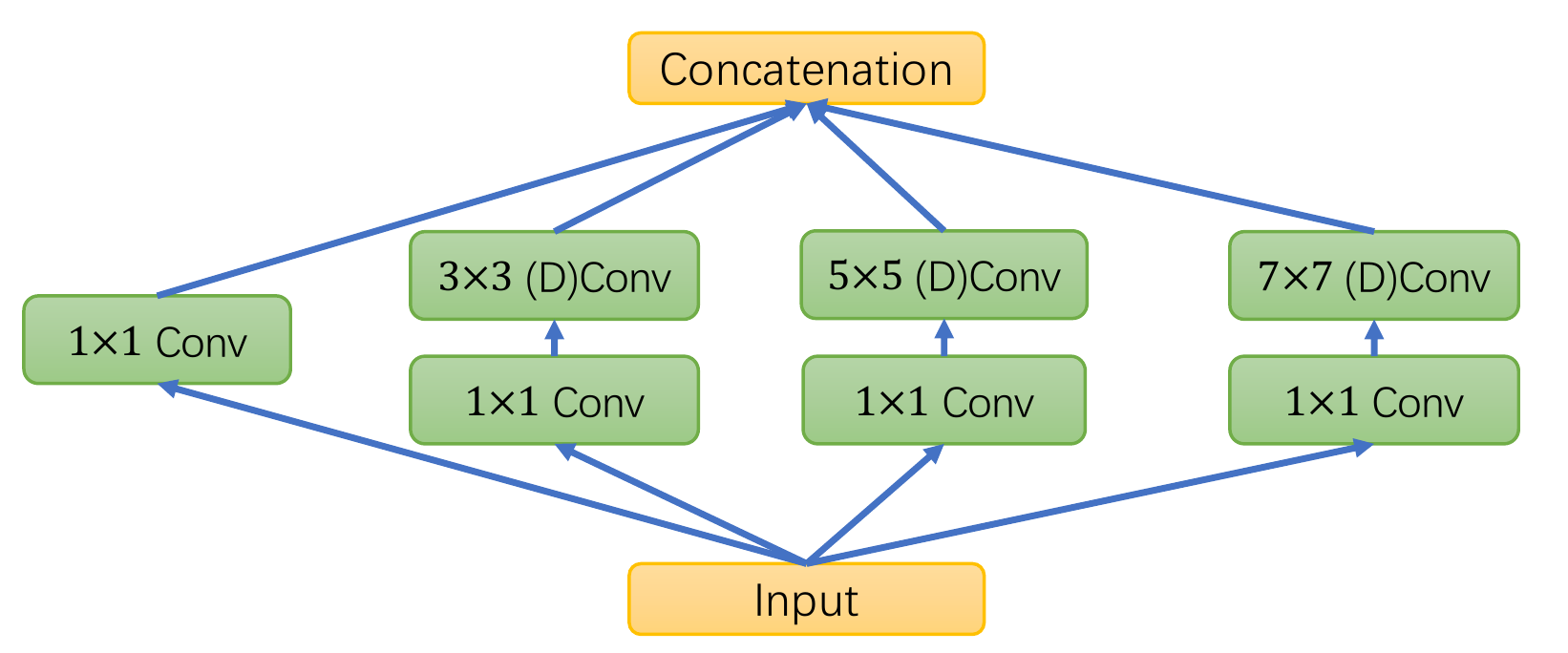}
	\caption{The architecture of the second module of LFE. \textbf{(D)Conv} represents the convolution layer with dilated kernels.}
	\label{fig:4}
\end{figure}

Specifically, LFE consist of two feature extractor blocks and each of them contains four branches with filter sizes of $1\times1$, $3\times3$, $5\times5$, and $7\times7$ respectively, as shown in Figure\ \ref{fig:4}. Each branch focuses on a certain scale and generates the same number of feature maps. To further enhance model's capability to capture multiple scales information, dilated convolution, which can enlarge the receptive field without involving extra computations, is employed in the second block. Additionally, expect for the $1\times1$ branch, there is an extra $1\times1$ filter added before the other three branches to reduce the feature channels by half. Moreover, ReLU activate function is applied after each convolution layer in LFE to avoid negative values.

As a departure from most of works, we remove the pooling layers between the inception modules to avoid the reduction in spatial resolution caused by the pooling operation and the additional complexity brought by subsequent deconvolutional layers. Considering the trade-off between resource consumption and model accuracy, we instead adopt dilated filters with the dilated rate of 2 to replace the pooling layer \cite{chen2018deeplab}. Features from different branches, covering multi-scale appearance of people in images, are subsequently concatenated together for the feature encoding.

\subsection{High-level Feature Encoder (HFE)}
The structure of HFE is shown in Figure\ \ref{HFE}, which takes shallow features extracted from the second block of LFE as input. While encoding features, such a structure can well preserve multi-scale information.

HFE is compose of two blocks, where each block consists of three convolution layers with the filter size of $3\times3$ followed by a ReLU activate function. Particularly, the input of a specific convolution layer $F_{l_j}$($ j=1,2,3 $) is the concatenation of all outputs from preceding layers, \ie, $F_{l_j} = concat(F_1,...,F_{l_{j-1}})$, which are indicated by different colors in the figure. The dense connection between layers ensures that multi-scale information in the shallow features can be preserved. At the bottom of each block, a $1\times1$ convolution layer is applied to integrate the concatenated hierarchical features and reduce feature channels to the same dimension as the input, which is indicated by $Conv_{1\times1}$. Therefore, the output of the $i$-th block in HFE $F_{L_i}$ can be obtained by,
\begin{equation}\label{equ:2}
F_{L_i}=F_{L_{i-1}}+Conv_{1\times1}[concat(F_{l_1},F_{l_2},F_{l_3})].
\end{equation}
Finally, the input of each block is added onto the output, which will in turn become the input of the next block.

On top of that, to further preserve multi-scale information, shallow features obtained by low-level feature extractor ($F_{M_2}$) and the output of each block in HFE ($F_{L_{i}}$) are concatenated together, which is $F_{g}$ in Eq. (\ref{equ:4}), as the input for the feature integration in global level. In the integration, a $1\times1$ and a $3\times3$ convolution layer are employed to integrate high-level features in a global level, which is indicated by $G$ in Eq. (\ref{equ:4}). Henceforth, the output of HFE can be calculated by,
\begin{equation}\label{equ:4}
F_G= G(F_{g}).
\end{equation}

Rather than widening the network, the proposed densely connected structure takes full advantage of features from all layers and well preserves the scale information in shallow features, which efficiently eliminates the problem of scale variation. In the paper, the dimension of $F_{M_2}$ and $F_G$ are both set to 64 according to the extensive experiments, which is less than most of the state-of-the-art methods. 

\subsection{Attention Map Generator (AMG)}
In light of the observation that backgrounds on shallow feature maps tend to have significantly different responses, compared to the crowds, we generate attention maps based on low-level features only. Specifically, AMG takes shallow features from the first block of LFE ($F_{M_1}$) as input and generates pixel-wise attention maps ($F_{att}$) on which crowd regions are always "brighter" than the backgrounds, \ie,
\begin{equation}\label{equ:att}
F_{att} = AMG(F_{M_1}).
\end{equation}

Here, two convolution layers followed by a sigmoid function, as shown in Figure\ \ref{AMG}, are used to ensure that all the computed weights are within the range of 0 to 1. $ L_{att} $, the summation of pixel-wise Euclidean distance between refined feature maps $ F_{ref} $ and ground-truth density map $ D $, conveys the supervision information to the learning process of the attention module. Subsequently, the attention map $F_{att}$ is employed to refine the encoded feature $F_G$ by element-wise multiply $(\otimes)$ as follows,
\begin{equation}\label{equ:5}
F_{ref} = F_{G} \otimes F_{att},
\end{equation}
where $F_{ref}$ is taken as the input of the last two convolution layers whose filter sizes are $1\times1$ and $3\times3$ respectively to generate the high-quality density map $\hat{D}$ under the supervision of a combination of several losses.

\begin{table}
	\centering
	\caption{Comparison results of different methods on the UCF\_CC\_50 dataset.}
	\label{tab:2}
	\begin{tabular}{|c|cc|}
		\hline
		Method & MAE & MSE \\  \hline
		FHSc+MRF & 468.0 & 590.3 \\
		MCNN  & 377.6 & 509.1 \\
		Switching-CNN  & 318.1 & 439.2 \\
		SANet \cite{cao2018scale} & 258.4 & 334.9 \\
		CSRNet \cite{li2018csrnet} & 266.1 & 397.5 \\
		SAAN \cite{hossain2019crowd} & 271.6 & 391.0 \\
		\SDANet{} (ours) & \textbf{227.6} & \textbf{316.4} \\
		\hline
	\end{tabular}
\end{table}

\begin{table*}
	\centering
	\caption{Comparison results of different methods on 5 scenes (S1$\sim$S5) in the WorldExpo$'$10 dataset in terms of MAE.}
	\label{tab:3}
	\begin{tabular}{|c|c|c|c|c|c|c|}
		\hline
		Method & S1 & S2 & S3 & S4 & S5 & Average\\
		\hline
		Cross-scene \cite{zhang2015cross} & 9.8	& 14.1 & 14.3 &	22.2 & 3.7 & 12.9 \\
		MCNN \cite{zhang2016single} & 3.4 & 20.6 & 12.9 & 13.0 & 8.1 & 11.6 \\
		Switching-CNN \cite{sam2017switching} & 4.4 & 15.7 & 10.0 & 11.0 & 5.9 & 9.4 \\
		SANet \cite{cao2018scale} & 2.6 & 13.2 & 9.0 & 	13.3 & 3.0 & 8.2 \\
		CSRNet \cite{li2018csrnet} & 2.9 & \textbf{11.5} & \textbf{8.6} & 16.6 & 3.4 & 8.6 \\
		SaCNN \cite{zhang2018crowd} & 2.6 & 13.5 & 10.6 & 12.5 & 3.3 & 8.5 \\
		\SDANet{} (ours) & \textbf{2.0} & 14.3 & 12.5 & \textbf{9.5} & \textbf{2.5} & \textbf{8.1}  \\
		\hline
	\end{tabular}
\end{table*}

\begin{figure*}
	\centering
	\includegraphics[width=\textwidth]{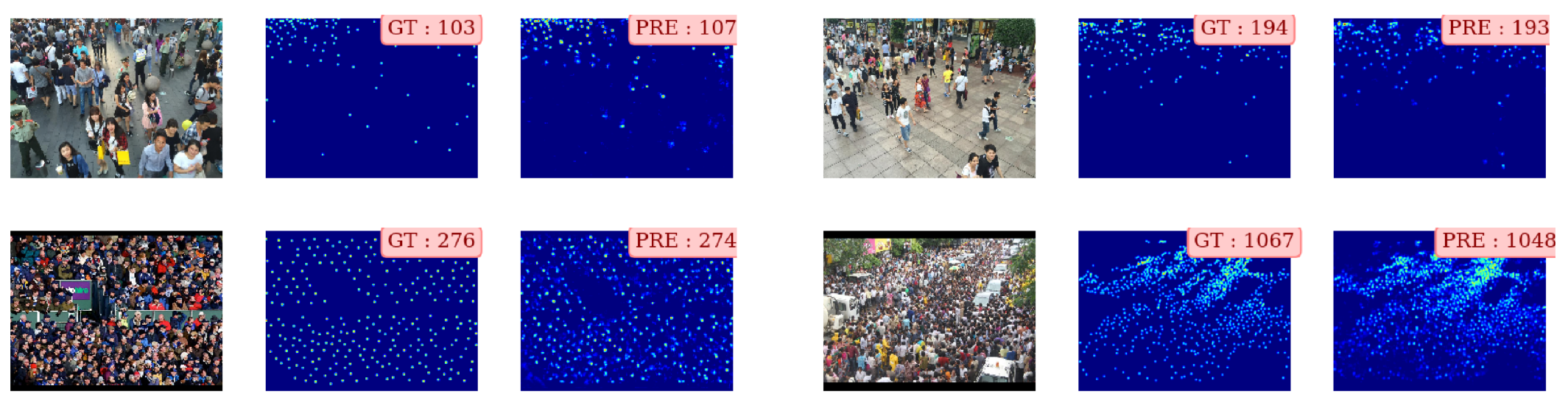}
	\caption{Qualitative results on ShanghaiTech Dataset. For each group of images, pictures in the middle and on the right are corresponding ground truth and estimated density map of the image on the left, where the number on the top right corner indicates the ground truth (GT) and the estimated number of people (PRE) respectively. It can be seen that \SDANet{} has a strong adaptability to different density levels with a error less than 4$\%$.}
	\label{fig:5}
\end{figure*}

\subsection{Loss Function}
The density maps generator in the \SDANet{} adopts a coarse-to-fine strategy. Concretely, the loss is composed of two terms: $L_{att}$ and $ L_{map}$ in the Figure\ \ref{SDANet} respectively. 

Firstly, a convolution layer with the filter size of $3\times3$ is employed to learn a coarse mapping between combined feature maps ($F_{ref}$) from the HFE and AMG to the density maps, and meanwhile, prepare coarse density maps for further process. In order to supervise the learning process of attention maps and the generation of coarse density maps, $L_{att}$, measuring the Euclidean distance between coarse density maps ($ \hat{D_C} $) and the ground-truth density map $D$, is adopted. Explicitly, $L_{att}$ is defined as,
\begin{equation}
L_{att} =\frac1{M} \sum_{M} ||\hat{D_C}-D||^2_2, 
\label{equ:7}
\end{equation}
where $M$ is the dimension of $\hat{D_C}$, and is set to 32 throughout all experiments.
 
Subsequently, two convolution layers with filter sizes of $3\times3$ and $1\times1$ are involved to further refine the quality of coarse density map, thus enhancing the accuracy of crowd counting. Noticeably, the ReLU activation function is employed after convolution layers to avoid appearance of negative values. Last, $ L_{map}$ is introduced to supervise the refinement process and generate the fine-grained density map ($\hat{D_F}$). Concretely, $L_{map}$ is composed by an Euclidean loss ($L_E$) and a Counting loss ($L_C$), which are somewhat complementary to each other. Initially, $L_E$ is adopted to improve the quality of density map by minimizing the Euclidean distance between the fine-grained density map and the ground-truth, which can be described by,
\begin{equation}
L_E =\frac1{N} \sum_{i=1}^{N} ||\hat{D_{F_i}}-{D_i}||^2_2, \label{equ:8}
\end{equation}
where $\hat{D_{F_i}}$ and $D_i$ are estimated density map and ground truth of the $i$-th image $I_i$, respectively, and $N$ refers to the number of training samples. However, sharp edges and outliers in coarse density maps might be blurry in fine-grained maps. To remedy this situation, $L_C$ is added as a compensation, which is defined by,
\begin{equation}
L_C = \frac1{N} \sum_{i=1}^{N}(\frac{C_i-\hat{C_i}}{C_i + \epsilon})^2, 
\label{equ:9}
\end{equation}
where $\hat{C_i}$ and $C_i$ represent, respectively, the estimated number of people and the ground truth of the $i$-th training sample, which are the integral over all pixels $ p $ on the corresponding density map, \ie, $ C_i=\sum_p D_i $. Additionally, $\epsilon=0.0001$ is set to avoid the denominator being zero. $L_C$ not only accelerates the convergence process but improves the counting accuracy. In summary, $L_{map}$ is expressed as,
\begin{equation}\label{equ:10}
L_{map} = L_{E} + \alpha L_{C},
\end{equation}
where $\alpha=0.01$ is the empirical weight for $L_C$. 

Therefore, the overall loss of \SDANet{} is,
\begin{equation}\label{equ:11}
Loss = L_{att} + L_{map}.
\end{equation}
Adam \cite{kingma2014adam} algorithm with the initial learning rate of 1e-4 is adopted to optimize the \SDANet{}.

\section{Experiments}
\subsection{Evaluation Metrics}

Similar to the previous work, the mean absolute error (MAE) and mean squared error (MSE) metrics are used for algorithm evaluation, which are defined as:
\begin{equation}
MAE = \frac{1}{N} \sum_{i=1}^{N} \left| C_i - \hat{C_i} \right|,\label{MAE}
\end{equation}
\begin{equation}
MSE = \sqrt[]{\frac{1}{N} \sum_{i=1}^{N} \left( C_i - \hat{C_i} \right)^2},\label{MSE}
\end{equation}
where $N$ represents the total number of images involved in testing, $C_i$ and $\hat{C_i}$ are the ground truth and estimated number of people for the $i$-th image respectively.

\begin{table*}
	\centering
	\caption{Comparison results of different methods on the ShanghaiTech dataset.}
	\label{tab:4}
	\begin{tabular}{|c|c|c|c|c|}
		\hline
		\multirow{2}*{Method} & \multicolumn{2}{|c|}{Part$\_$A} & \multicolumn{2}{|c|}{Part$\_$B} \\
		\cline{2-5}
		~ & MAE & MSE & MAE & MSE \\
		\hline
		Cross-scene & 181.8 & 277.7 & 32.0 & 49.8 \\
		MCNN & 110.2 & 173.2 & 26.4 & 41.3 \\
		Switching-CNN  & 90.4 & 135.0 & 21.6 & 33.4 \\
		CP-CNN \cite{sindagi2017generating} & 73.6 & 106.4 & 20.1 & 30.1 \\
		DecideNet \cite{liu2018decidenet} & - & - & 21.5 & 32.0 \\
		ACSCP \cite{shen2018crowd} & 75.7 & 102.7 & 17.2 & 27.4 \\
		CSRNet & 68.2 & 115.0 & 10.6 & 16.0 \\
		SANet  & 67.0 & 104.5 & 8.4 & 13.6 \\
		TEDnet \cite{jiang2019crowd} & 64.2 & 109.1 & 8.2 & 12.8 \\
		\SDANet{} (ours) & \textbf{63.6} & \textbf{101.8} & \textbf{7.8} & \textbf{10.2} \\
		\hline
	\end{tabular}
\end{table*}

\subsection{Datasets}
In the experiment, three crowd counting benchmark datasets, the UCF\_CC\_50 dataset, the WorldExpo$ ' $10 dataset, and the ShanghaiTech dataset, are used to evaluate the performance of \SDANet{}, each being elaborated below. 

\subsubsection{UCF\_CC\_50 dataset}
\cite{idrees2013multi} contains 50 images with various perspectives and resolutions. The number of annotated people per image ranges from 94 to 4543 with an average number of 1280, which is a challenging dataset in the field of crowd counting.

\subsubsection{WorldExpo$ ' $10 dataset}
\cite{zhang2015cross} consists of 3980 annotated frames from 1132 video sequences captured by 108 different surveillance cameras, which is divided into a training set (3380 frames) and a test set (600 frames). The region of interest (ROI) is also provided for the whole dataset. 

\subsubsection{ShanghaiTech dataset}
\cite{zhang2016single} consists of 1198 annotated images with a total amount of 330,165 annotated people. The dataset contains two parts: Part$\_$A and Part$\_$B. Part$\_$A includes 482 internet images with highly congested scenes while Part$\_$B includes 716 images with relatively sparse crowd scenes taken from streets in Shanghai. 

\begin{table}
	\centering
	\caption{Ablation study results on the WorldExpo$'$10 dataset.}
	\label{tab:5}
	\begin{tabular}{|c|cc|}
		\hline
		Models & MAE & MSE \\
		\hline
		\SDANet{} without AMG & 12.89 & 15.28 \\
		\SDANet{} without Dense Structure & 10.14 & 13.25 \\
		\SDANet{} without Refinement & 9.64 & 13.19  \\
		\SDANet{} & 8.10 & 12.90 \\
		\hline
	\end{tabular}
\end{table}

\subsection{Experiment Settings}
Taking the computation cost and data variety into account, we adopted the patch-wise training strategy. Following the previous work \cite{zhang2016single}, 9 patches, where each patch is $1/4$ of the image size, are cropped from each image to generate the training set. The first four patches contain four quarters of the image without overlapping while the other five patches are randomly cropped from the image. During the test, non-overlapping patches are cropped from each image in the test set and compute individually. The final density map of the image is the concatenation of its patches’ predictions. Additionally, images are further augmented by randomly horizontal flipping.

Besides, we generated the ground-truth from head annotations given by datasets \cite{zhang2016single}. Each head annotation is blurred with a Gaussian
kernel, whose summation is normalized to one and the number of people is the integral over the density map. 

The implementation of \SDANet{} is based on the PyTorch framework. As we train the whole network from scratch, all parameters are randomly initialized by Gaussian distribution with mean of zero and standard deviation of 0.01.

\section{Results and Analysis}
On each dataset, we follow the standard protocol to generate ground truth and compare our method with the state-of-the-art algorithms. Furthermore, we conduct extensive ablation experiments on the WorldExpo$ ' $10 dataset to analyze the effects of different components in \SDANet{}. We explain experimental settings and show results as follows.

\subsection{Experimental Evaluations}
\subsubsection{Quantitative results}
On the \textbf{UCF\_CC\_50} dataset, we performed a 5-fold cross-validation to evaluate the proposed method as suggested by \cite{idrees2013multi}. Table. \ref{tab:2} shows the comparison of the results of our method with contemporary state-of-the-art works on UCF\_CC\_50 dataset, which illustrates the proposed \SDANet{} is able to deal with crowd scenes with varying densities and achieves a superior performance over other approaches. Specifically, our method achieves 11.91$\%$ MAE reduction and 5.52$\%$ MSE reduction. This clearly demonstrates that \SDANet{} is super robust against the scale and density changes. 

The comparison results of \SDANet{} with contemporary state-of-the-art work on the 5 scenes (S1$\sim$S5) in the test set of \textbf{WorldExpo$ ' $10} dataset are shown in Table. \ref{tab:3}. The challenging test set is a combination of different densities, ranging from sparse to dense, and various backgrounds including squares, stations, \etc. From the result, it can be seen that the proposed \SDANet{} scores the best in Scene1, Scene4 and Scene5 as well as the best accuracy on average, which again proves the strong adaptability of \SDANet{} against different scenarios with varying density levels.

On the \textbf{ShanghaiTech} dataset, \SDANet{} is evaluated and compared with other recent works and results are shown in Table. \ref{tab:4}. Again, the proposed method attains the lowest MAE and MSE as well. Specifically, our approach outperforms the latest work TEDnet by 4.87$\%$ and 20.31$\%$ over the MAE and MSE metric respectively on the ShanghaiTech Part$\_$B dataset. 

\subsubsection{Visualization results} 
We firstly analyzed the attention maps generated by AMG and obtained some statistical results. Taking the attention map of Figure\ \ref{MCNN.1} as an example, the average attention value of crowd region (center-right) is 0.874 (GT=1) while that for background region (left corner) is 0.253 (GT=0), which proves that the attention maps reduce the background noise by arranging background regions with relatively low weights. 

To demonstrate the performance of \SDANet{} on scenes with cluttered backgrounds and varying head sizes, we choose, in particular, the ShanghaiTech dataset for estimated density maps visualization, which are shown in Figure\ \ref{fig:5}. For each group of images, pictures in the middle and on the right are corresponding ground truth and estimated density map of the image on the left, where the number on the top right corner indicates the ground truth (GT) and the estimated number of people (PRE) respectively. Here, we display the estimated density maps of various scenarios, ranging from 103 persons to 1067 persons, to demonstrate that the proposed \SDANet{} performs decently in both dense and sparse scenes. It can be seen that \SDANet{} has a strong adaptability to different density levels with a error less than 4$\%$.

\subsection{Ablation Study}
To validate the effectiveness of key components in the \SDANet{}, we also conducted ablation studies on the WorldExpo$'$10 dataset which is more realistic and challenging due to the fact that all images are acquired from real surveillance scenes.
 
\subsubsection{Effectiveness of AMG} We explore the performance improvement offered by AMG by removing the attention module from the \SDANet{} and compare it with the network with AMG. The result is indicated by \textit{\SDANet{} without AMG} in Table. \ref{tab:5}. There are 37$\%$ increase in MAE and 15$\%$ increase in MSE if AMG  is dropped out, clearly demonstrating that AMG has made a significant contribution in diminishing background noise.

\subsubsection{Effectiveness of densely-connected structure} In order to shed light on how the densely connecting structure preserves multi-scale features, we conduct an experiment on the same dataset without the dense connection between layers and the result is indicated by \textit{\SDANet{} without Dense Structure} in Table. \ref{tab:5}. It can be seen that the removal of the dense connection between layers leads to an over 20.1$\%$ drop in the counting accuracy, which means that densely-connected structure reinforces the diversity of features and improve the performance of \SDANet{}.
 
\subsubsection{Effectiveness of estimation refined layers} Furthermore, we study the refinement ability of the last two layers and the loss term $L_{map}$. We screen out the last two convolution layers in \SDANet{} and train the network with solely $L_{att}$, whose result is indicated by \textit{\SDANet{} without Refinement} in Table. \ref{tab:5}. Without the refinement layers, there is a nearly 16$\%$ decline in the MAE. Therefore, the coarse-to-fine strategy involved in the loss function can further enhance the performance of the network.

\section{Conclusion}
In this paper, we have presented a brand-new Shallow feature based Dense Attention Network (\SDANet{}) aiming to automatically count the number of people in an image. Our \SDANet{} is characterized by: 1) diminishing the impact of backgrounds via involving a lightweight attention model, and 2) capturing multi-scale information via densely connecting hierarchical image features. Extensive experiments have been carried out and the results on three benchmark datasets validate the adaptability and robustness of the \SDANet{} when varying crowd scenes from sparse to dense. 

\bibliographystyle{aaai} 
\bibliography{ref}

\end{document}